\documentclass[letterpaper, 10 pt, conference]{ieeeconf-modified}  %

\IEEEoverridecommandlockouts                              %

\usepackage[numbers]{natbib}
\usepackage{graphicx}
\graphicspath{ {./figures} }
\usepackage{siunitx}
\usepackage{algorithm}
\usepackage{algorithmic}
\usepackage{amsmath} %
\usepackage{amssymb}  %
\usepackage{textcomp}
\usepackage{hyperref}
\usepackage[capitalize]{cleveref}
\usepackage{capt-of}

\DeclareMathOperator{\clip}{clip}

\crefname{equation}{}{}

\title{\LARGE \bf
Learning Controlled Separation of Small Objects \\Between Two Fingers with a Tactile Skin
}
\author{Ulf Kasolowsky \;\;\;\;\;  Berthold Bäuml%
}

\begin{document}

\twocolumn[{%
        \renewcommand\twocolumn[1][]{#1}%
        \maketitle
        \begin{center}
            \centering
            \includegraphics[width=\textwidth]{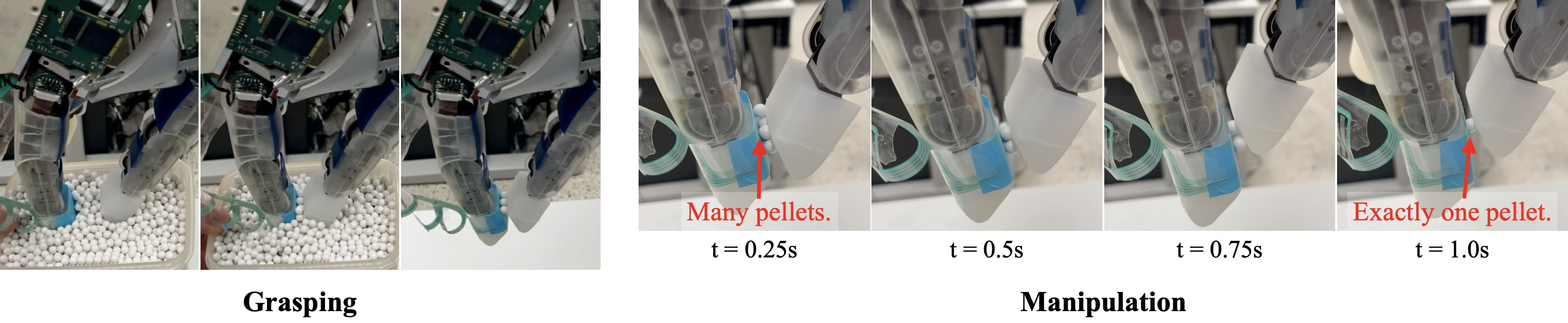}
            \captionof{figure}{
                \textbf{Left:} Grasping procedure. 
                The two fingers are in an open configuration and "dive" into the box with the pellets. 
                After closing the fingers, the box is removed and multiple pellets remain between the fingertips.
                \textbf{Right:} Manipulation procedure.
                The policy moves the fingers so that pellets drop until only the desired number, in this case one ($P_d=1$), remains between the fingers (see also the accompanying video; \cref{fig:sequence_3} shows the case with $P_d=3$).
            }
            \label{fig:title_figure}
        \end{center}%
    }]

{\let\thefootnote\relax\footnote[0]{
   \noindent The authors are with the Learning AI for Dextrous Robots Lab \href{https://aidx-lab.org}{(aidx-lab.org)}, Technical University of Munich, Germany, and the DLR Institute of Robotics and Mechatronics, German Aerospace Center.\newline
Email:{\tt\scriptsize{ \{ulf.kasolowsky|berthold.baeuml\}@tum.de}}\newline}}

\thispagestyle{empty}
\pagestyle{empty}

\begin{abstract}
We introduce and solve the novel task of \emph{controlled separation} of small objects with two fingers of a multi-purpose robotic hand: after grasping into a box of small objects, the task is to drop as many of them until a desired number remains between the fingers. The objects are small compared to the width of the fingers but also in absolute terms. In our case little pellets with a diameter of only \SI{6}{\milli\meter} are handled.
We show that the task can be performed purely tactile (no vision) using a spatially-resolved tactile skin on a fingertip.
The separation policy is trained in simulation via reinforcement learning using a straightforward sparse reward, which basically checks if the desired number of objects is reached.
In simulation experiments, we provide an exhaustive analysis of the benefits of using spatially-resolved tactile feedback:
while an ideal (high-resolution) tactile sensor allows solving the task almost perfectly, a sensor with lower spatial resolution (here $4\times 4$ taxels) still leads to an improvement of up to \SI{20}{\percent} compared to using only the fingers' joint sensors.
For this analysis, we further train an estimator alongside the policy that predicts the ground truth contact positions.
Finally, we demonstrate the successful sim-to-real transfer for the DLR-Hand II equipped with a tactile skin.\\
Website: \href{https://aidx-lab.org/skin/icra26}{aidx-lab.org/skin/icra26}
\end{abstract}

\section{Introduction}
Blind bin picking of small objects is an important skill, e.g. for many industrial tasks.
Humans typically solve this by grasping with two fingers into a box full of small objects and taking out what is (by chance) kept between the fingertips. If there are too many objects, some can be dropped in a controlled manner by moving the fingers until the desired number of objects is reached.
We refer to this task as \emph{controlled separation}.
To complete this task, humans are typically relying on their spatially-resolved sense of touch.

In this paper, we introduce and solve the task of controlled separation of small pellets between two fingers of a multi-purpose robotic hand (see \cref{fig:title_figure}).
The pellets have a diameter of \SI{6}{\milli\meter} and could, for example, be medical products or the balls of a small bearing.
To provide a sense of touch, we add a tactile skin to one of the fingertips.
To solve the task, we use deep reinforcement learning in simulation and then transfer the the resulting policy to the real system.

Such a fine manipulation skill significantly increases the range of applications of multi-fingered robotic hands. 
Such hands have only recently been enabled by advanced reinforcement learning (RL) methods and training in simulation to perform tasks like grasping arbitrary unknown objects or in-hand manipulation -- coming ever closer to human dexterity.

\subsection{Related Work}
To the best of our knowledge, the separation of objects which are small compared to the size of the gripper or the finger width has so far not been studied for a multi-purpose robotic hand.

In \citet{zhao2022}, a parallel-jaw gripper with a length-adjustable jaw is used to bin pick a single object.
For this, they train a grasp-prediction network that receives a depth image as input.
However, there is no separation happening after the grasp and the objects are not small but have a size similar to the width of the gripper. 
\citet{ohara2025} present a custom gripper to pick exactly one M3 or M6 screw from a bin.
After grasping into the bin, they perform a shaking motion and estimate the number of grasped objects based on the resulting frequency pattern.
In case more than one object was picked up, the process is repeated. In contrast to our work, there is no separation strategy.
In \cite{zhou2024,zhou2025}, another custom gripper is used to grasp an object out of a granular medium.
Afterwards, a rubbing motion is performed to separate the object from pieces of granular medium.
This task, however, is significantly easier as the grasped objects are way bigger compared to the medium itself.

\citet{do2024} perform inter-finger manipulation of objects down to \SI{5}{\milli\metre} size.
They grasp the objects from a bowl but do not focus on the separation but only on the reorientation for a downstream classification task.
The authors explicitly mention that it is challenging to solve the manipulation of such small objects with reinforcement learning because it is hard to simulate the dynamics of the soft fingertips correctly.
Instead, they rely on a handcrafted control algorithm.

\citet{ishige2020blind} present the separation of small screws with a specialized two-finger gripper with 3 DOF which is equipped with a spatially-resolved tactile sensor. 
They use  reinforcement learning on the real system and the goal is to reduce the number of screws to one. 
Although this work is the closest one to our setting, their task is significantly simpler. 
The task to reduce the number to one instead of to a given desired number is fundamentally easier so that their resulting policy is basically acting randomly. 
In addition, they use a special gripper with only 3 DOF, whereas our policy has to control the 6 DOF of the two fingers of a multi-purpose hand. 
Finally, our policy is trained in simulation with robust sim-to-real transfer, massively reducing the training effort compared to learning directly on the real system.

\subsection{Contributions}
Our main contributions are as follows:
\begin{itemize}
    \item We present the novel task of \emph{controlled separation} of small objects, i.e., a desired goal number $\ge 1$ of objects has to remain between the fingers.
    \item We solve this task using reinforcement learning in simulation using only a simple natural and sparse reward. 
    \item We provide a detailed analysis in simulation demonstrating the effectiveness of tactile feedback for this task.
    We compare the tactile skin that is available on the real robot with a ground truth encoding of the pellet contact positions.
    \item We perform experiments on the real DLR-Hand II  with a tactile skin proving the successful sim-to-real transfer.
\end{itemize}

\section{Task Definition and Setup}
First, the robot grasps blindly into a box of pellets.
After that, the goal is to separate the pellets, i.e. dropping some in a controlled manner, until a certain number $P_d$ remains in a stable grasp.
The pellets could then be used for downstream tasks like packaging or assembly.

For the initial grasp into the box, we rely on a manual strategy illustrated on the left side of \cref{fig:title_figure}.
To pick up as many pellets as possible, we plan an initial joint configuration $q_0$ where the flat surfaces of the thumb and forefinger are as parallel as possible to each other.
On the real system, the robot first digs into the box of pellets with slightly opened fingers.
The fingers then approach the closed position $q_0$ before each applying a force of $\SI{1.5}{\newton}$ in the grasp normal direction.

The separation strategy is learned in simulation using deep reinforcement learning and then transferred to the real robotic system. 
In this paper, we test the separation to one to three pellets: $P_d \in \{1,\,2,\,3\}$.
An example run is shown on the right side of \cref{fig:title_figure} for $P_d=1$.

\begin{figure}[htb]
    \centering
    \includegraphics[width=1.0\linewidth]{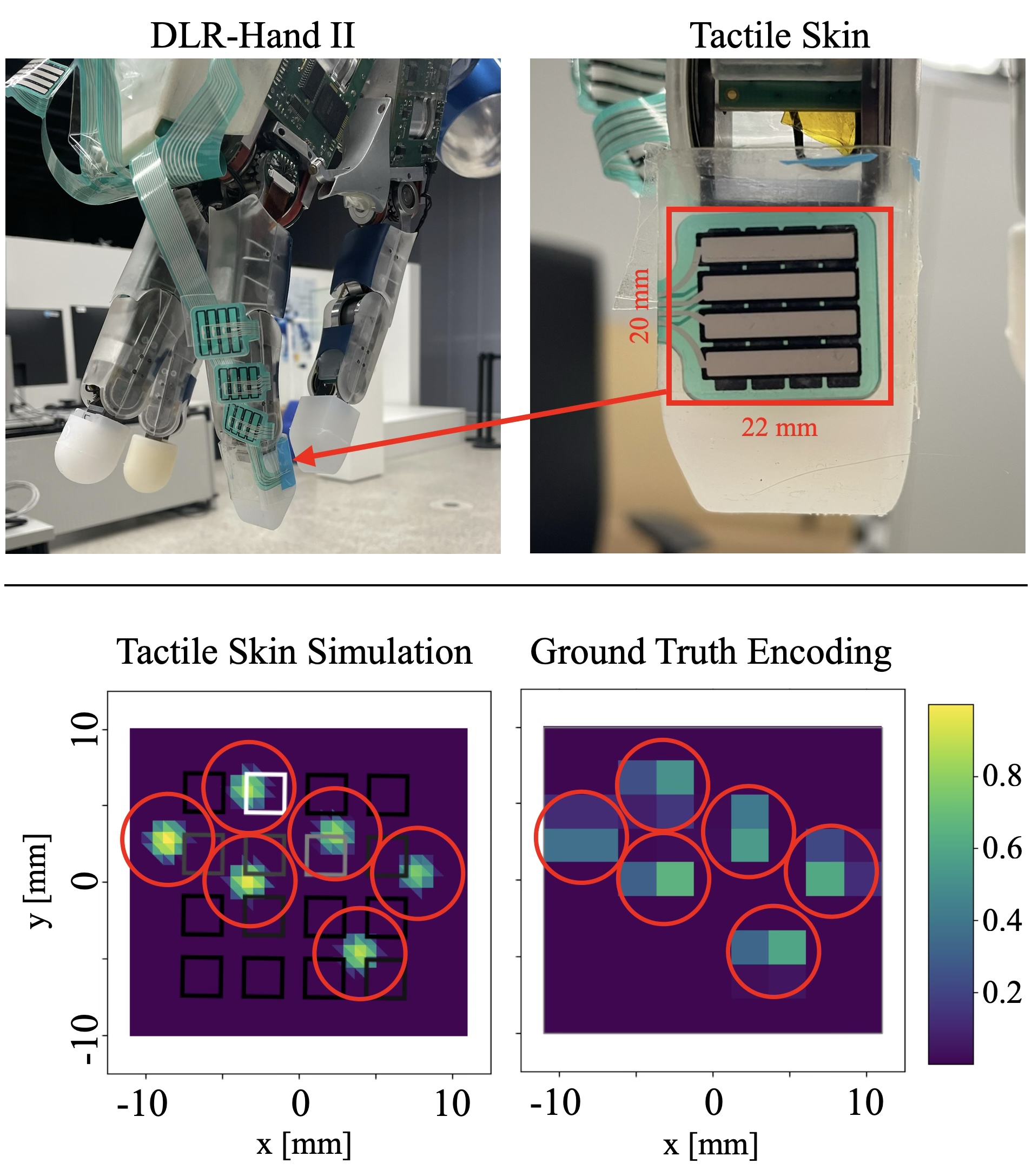}
    \caption{
        \textbf{Top:} Hardware setup. 
        We use the DLR-Hand II together with a tactile skin. 
        Onto the tactile skin we add a small layer of rubber (visible in blue on the left) for better friction.
        \textbf{Bottom:} Tactile sensor simulation.
        The red circles indicate the pellets on the fingertip.
        On the left, the simulation of the real tactile sensor with $4 \times 4$ taxels is shown.
        The shown heatmap visualizes the normalized simulated pressure distribution. 
        The squares represent the individual taxels, whose response is calculated by integrating the pressure over their area (black: no activation, white: high activation).
        On the right, the ground truth encoding of the contact positions is visualized.
        Each rectangle of the heat map represents a probability that a pellets lies within it.
    }
    \label{fig:tactile_sensor}
\subsubsection*{Simulation}
\end{figure}

We perform the experiments on the DLR-Hand II \cite{butterfass2001} of the humanoid robot Agile Justin \cite{bauml2014} shown in the top left of \cref{fig:tactile_sensor}.
In addition to the proprioceptive feedback of the joints, we attach a TekScan sensor (top right of \cref{fig:tactile_sensor}) to the forefinger to provide spatially-resolved tactile feedback.
The hardware setup and its simulation are explained in more detail in the following.

\subsection{DLR-Hand II}
The DLR-Hand II is a multi-purpose robotic hand.
It has been used successfully for both dextrous in-hand manipulation of objects up to \SI{20}{\centi\metre} \cite{rostel2025} and fine manipulation of small marbles with a radius as small as \SI{4}{\milli\metre} \cite{kasolowsky2024}.
The DLR-Hand II has four fingers with three active degrees of freedom each.
Each joint is controlled with a joint impedance controller running at \SI{1000}{\hertz}.

To learn the separation task in simulation, we need a model of the hand. 
We use the Mujoco rigid-body simulator \cite{todorov2012} to implement a model including the joint impedance controller similar to our previous work \cite{sievers2022, kasolowsky2024}.
To save computation costs, we only simulate the two fingers in use.
Moreover, the simulator runs at only \SI{200}{\hertz} compared to the real-life control rate of \SI{1000}{\hertz}.

\subsection{Tactile Skin}
As a tactile skin we use a piezoresistive pressure sensor by TekScan, which can be simply glued to the fingertip (see also the top row in \cref{fig:tactile_sensor}).
For better friction, we attach a rubber glove on top.
The sensor consists of a $4 \times 4$ measurement array.
We refer to the individual measurement cells as taxels in the following.
Each taxel has a size of $\SI{2.5}{\milli\metre}\times\SI{2.5}{\milli\metre}$ with \SI{1.5}{\milli\metre} gaps in between.
This leads to a total dimension of $\SI{14.5}{\milli\metre}\times\SI{14.5}{\milli\metre}$.
Note that the flat surface of the fingertip has a dimension of $\SI{22}{\milli\metre}\times\SI{20}{\milli\metre}$, i.e., the tactile skin is blind at the edges.

\subsubsection*{Simulation}
To simulate the tactile skin, we use a modified version of the tactile simulation presented in \cite{kasolowsky2024}.
A sample of the skin model is visualized on the bottom left of \cref{fig:tactile_sensor}.
First, the correct pressure distribution of each pellet pressing onto the fingertip is calculated based on the normal force, contact normal, and relative positions obtained from the Mujoco simulation.
As can be seen in the visualization, the pressure distribution is represented on a mesh.
To obtain the tactile image $T$, which is also provided by the real sensor, for each taxel $T_i$, the pressure inside its area is integrated giving a resulting force acting on that particular taxel.
As the tactile images are computed based on the state of the physics simulator but do not directly feed back to it, it is sufficient to evaluate the skin model only when a new observation for the reinforcement learning policy is required.

\subsubsection*{Ground Truth Encoding}
\label{sec:gt_encoding}
As stated above, the tactile sensor used in this paper is not ideal in the sense that it has a very low spatial resolution and does not cover the entire fingertip.
To analyze how well the task could be performed with a "perfect" sensor, we test training the policy with a ground truth encoding $T^*$ of the pellet positions.
For this encoding, we first create a uniform grid of points $t_i$ on the entire flat part of the fingertip with a resolution of $9 \times 9$.
Then for every pellet contact position projected onto the flat part of the fingertip $p_j$, the distance to these points can be computed in x and y direction:
\begin{equation}
    d_{i,j,x/y} = \left|p_{j,x/y} - t_{i,x/y}\right|,\quad j \in P_c,
\end{equation}
where $P_c$ is the current number of pellets in contact.
The encoding value for each point $t_i$ and each pellet $p_j$ is then computed as:
\begin{equation}
    T_{i,j}^* = \max\left(1-\frac{d_{x,i,j}}{d_{x,\mathrm{max}}},\,0\right) \cdot \max\left(1-\frac{d_{y,i,j}}{d_{y,\mathrm{max}}},\,0\right),
\end{equation}
where $d_{x/y,\mathrm{max}}$ are the distances between two neighboring points in $x$ and $y$ direction, respectively. 
This encoding $T_{i,j}^*\in[0,\,1]$ exactly defines the pellet position and can be interpreted as a probability that the contact point is in a certain area.
The resolution is chosen in a way that the encodings are non-overlapping for the different pellets:
\begin{equation}
    T_{i,k} \neq 0 \implies T_{i,l}=0 \quad \forall l \neq k.
\end{equation}
Therefore, the pellet positions can still be uniquely reconstructed from the combined encoding $T_i^*=\sum_j T_{i,j}^*$. 
A sample of the ground truth encoding $T^*$ is shown on the bottom right of \cref{fig:tactile_sensor}.

\section{Learning Purely Tactile Separation}
We want to learn the task in simulation using the PPO algorithm \cite{schulman2017}.
In this section, we present the environment and learning setup.

\subsection{Initial Grasp Distribution}
\begin{figure}
    \centering
    \includegraphics[width=1.0\linewidth]{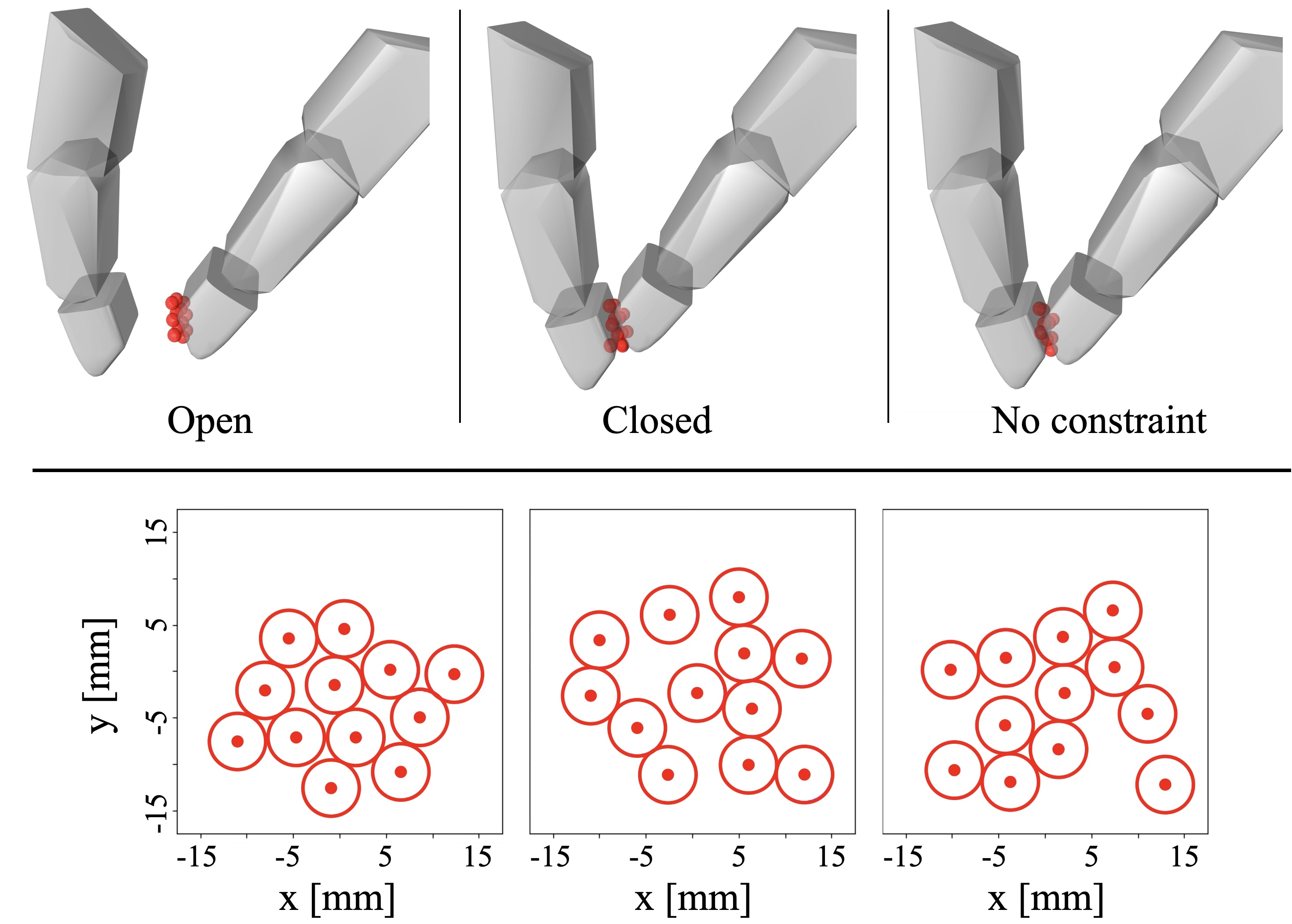}
    \caption{\textbf{Top:} Grasping procedure in simulation. 
    Pellets get spawned between the opened fingers in a random initial configuration and are held in place via a constraint.
    The fingers then close and the constraint is removed. 
    Only pellets that are stably grasped remain between the fingertips, the other ones drop due to gravity.
    \textbf{Bottom:} Three samples of initial pellet configurations. 
    The distribution approximates how pellets remain between the closed fingers in reality.}
    \label{fig:sim_distribution}
\end{figure}

In simulation, we do not simulate the entire box of pellets. 
Simulating more than \num{500} pellets in a rigid-body simulator is unfeasible as it requires very low time steps to keep the simulation stable.
Instead, we spawn a random configuration of twelve pellets on the two-dimensional contact plane between the two fingertips.
Initially, the fingers are again in a slightly opened position and the pellets are held in place via a constraint.
As on the real system, the fingers then close and start pressing on the pellets. 
Now the constraint is removed and the free-floating pellets drop due to gravity while the grasped ones remain between the fingers.
The procedure is visualized at the top of \cref{fig:sim_distribution}. 
Note the similarity of the final state compared to the real one in \cref{fig:title_figure}, showing that it is not necessary to simulate the entire box of pellets.

To simulate the distribution of the pellets realistically, we initialize their position on the contact plane according to the close-packing of equal spheres.
The resulting arrangement is randomly shifted and rotated. 
Eventually, Gaussian noise is added to the pellet positions, and potential overlaps are resolved.
See the bottom of \cref{fig:sim_distribution} for samples of the resulting distribution.
To imitate the impact of surrounding pellets on the joint movement while closing, we add random offsets to the configuration $q_0$ before closing the finger.
With this procedure we generate $\SI{6600}{}$ initial states while rejecting the ones with less than six stably grasped pellets.

\subsection{Learning Objective}
\subsubsection*{Reward}
To formalize the task into a reward function used for reinforcement learning, we stick to first principles without highly specialized reward tuning.
The policy receives a small positive reward if the number of pellets in contact with the forefinger $P_c$ equals the desired one $P_d$. 
This motivates the policy to reach the target number of pellets as quickly as possible by dropping the superfluous ones.
To incentivize a stable grasp, we only grant this reward if all contact points lie on the front side of the fingertip. 
This prevents the policy from rolling the pellets on top of the tips or clamping them on the sides. 
In total, the reward is given as:
\begin{align}
    \begin{split}  
    r &= c_1  c_2, \\
    c_1 &= \left\{
        \begin{array}{ll} 
            1, & P_c = P_d \\
            0, & P_c \neq P_d
        \end{array}\right., \\
    c_2 &= \left\{
        \begin{array}{ll} 
            1, & \text{all contact points on foreside} \\
            0, & \text{else}
        \end{array}\right..
    \end{split}
\end{align}

\subsubsection*{Termination}
Once the policy has dropped too many pellets, it cannot recover from this state again so it does not receive any signals useful for learning anymore. 
Hence, we terminate the environment once too many pellets are dropped. 
Moreover, to avoid spending time in states where the correct number is reached already and to get more diverse data, we time out each environment after $\SI{5}{\second}$.

\subsection{Architecture}
An overview of the entire control architecture is given in \cref{fig:architecture}.
In the following, the individual aspects are explained in more detail.
\subsubsection*{Action Space}
The impedance controller of the DLR-Hand II receives a desired joint configuration $q_d \in \mathbb{R}^6$ as input for the thumb and forefinger. 
We use the output of the policy network to command a joint velocity $\Delta q$.
Therefore, the original actions $a\in[-1;\; 1]$ are scaled and added to the previous desired configuration. 
Moreover, we clip the desired joint angles at the joint limits:
\begin{equation}
    \tilde{q}_d^k = \clip\left(q_d^{k-1} + \Delta q,\, q_\text{min},\, q_\text{max}\right),\quad \Delta q = a \frac{\Dot{q}_\text{max}}{f_\text{cont}}, 
\end{equation}
where $q_\text{min/max}$ are the joint limits, $\Dot{q}_\text{max} = \SI{0.4}{\radian\per\second}$ is the maximum allowed joint velocity for this task, and $f_\text{cont}=\SI{20}{\hertz}$ is the control rate of the policy.
To further prevent damage to the hardware, we enforce a torque limit by clipping the difference between the desired and measured joint angles, eventually yielding the final output that is passed to the impedance controller:
\begin{equation}
    q_d = \clip\left(\tilde{q}_d,\, q_m-\frac{\tau_\text{max}}{K},\,q_m+\frac{\tau_\text{max}}{K}\right),
\end{equation}
where $\tau_\text{max}=\SI{0.5}{\newton\metre}$ is the maximum allowed torque.

\subsubsection*{Observation Space}
\begin{figure*}
    \centering
    \includegraphics[width=0.9\linewidth]{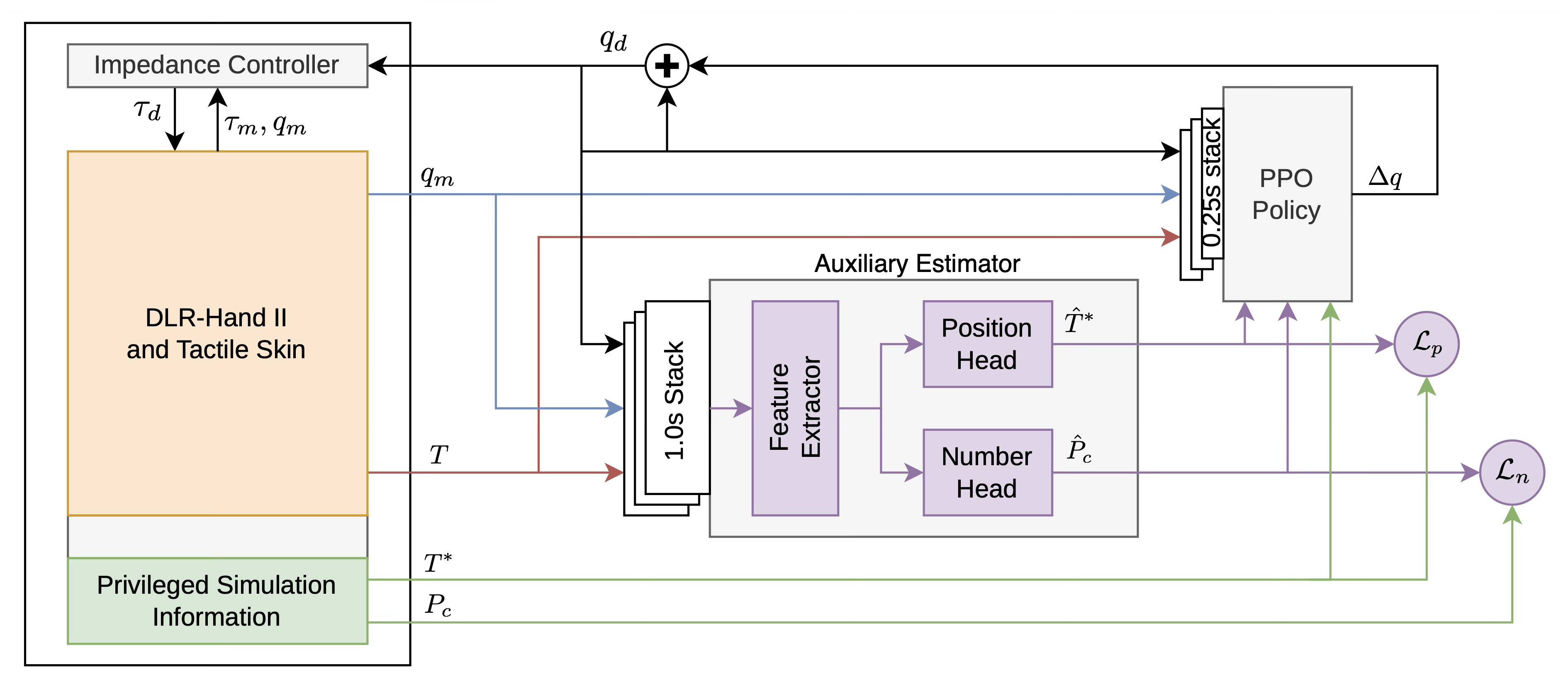}
    \caption{
        Control architecture. 
        On the left side, the controlled system is depicted, which can either be the real robot or the simulation. 
        It includes the hand with its joint impedance controller and the tactile skin.
        The policy, shown in the top right, outputs a joint velocity $\Delta q$ that is integrated to a desired joint angle $q_d$ and passed to the joint impedance controller (black signal).
        Together with the measured joint angles $q_m$ (blue signal), it is also fed back to the observation stack of the policy.
        Additionally, the tactile skin measurements $T$ (red signal) can also be fed into this observation stack.
        In simulation, we test passing further input to the policy, such as the ground truth encoding of the pellet positions $T^*$.
        Privileged information that is only available in simulation is shown in green.
        We also test passing the prediction of an auxiliary estimator (purple signals), shown in the middle, to the policy.
        The estimator has a bigger observation stack compared to the policy but also receives a combination of the latest desired joint angles $q_d$, measured joint angles $q_m$, and tactile images $T$.
        The estimator tries to predict the pellet positions $\hat{T}^*$ and current number of pellets $\hat{P}_c$.
        It is trained in a supervised manner alongside the policy using privileged simulation information $T^*$ and $P_c$ with the loss functions $\mathcal{L}_p$ and $\mathcal{L}_n$.
    }
    \label{fig:architecture}
\end{figure*}
We try a variety of different observation spaces to learn and analyze the task.
An overview can be found in \cref{tab:observation_spaces}.
First, we test only passing the joint angles to the policy without any tactile information (policy J).
Secondly, we provide additional tactile information via the simulated tactile skin (policy T) and via the ground truth encoding of the pellet positions (policy $\text{T}^*$).
Eventually, we further add the output of the auxiliary estimators explained in \cref{subsec:auxiliary_estimator} to the joint angle and tactile skin based observation spaces (policies $\hat{\text{J}}$ and $\hat{\text{T}}$).

As also shown in \cref{fig:architecture}, we use observation stacking to enrich the input signal to the policy. 
The observations of the desired joint angles $q_d$, the measured joint angles $q_m$, and the tactile skin $T$ are stacked five times.
Hence, the policy sees a history of the last $\SI{0.25}{\second}$.

\begin{table}[!htb]
    \caption{Observation spaces of different policies.}
    \label{tab:observation_spaces}
    \centering
    \begin{tabular}{lcccccc}
       
        \multicolumn{2}{c|}{Observation} & $\mathrm{J}$ & $\mathrm{T}$ & $\mathrm{T^*}$ & \multicolumn{1}{|c}{$\hat{\text{J}}$} & $\hat{\text{T}}$ \\\hline
        Desired Joint Angles & $q_d$ & $\checkmark$  & $\checkmark$  & $\checkmark$  & $\checkmark$  & $\checkmark$  \\
        Measured Joint Angles & $q_m$ & $\checkmark$ & $\checkmark$ & $\checkmark$  & $\checkmark$  & $\checkmark$ \\
        Tactile Image & $T$ &  & $\checkmark$ & & & $\checkmark$ \\
        Ground Truth Encoding & $T^*$ &   &  & $\checkmark$  \\
        Auxiliary Estimates &  $\hat{T}^*\text{, }\hat{P}_c$& & & &  $\checkmark$ & $\checkmark$
    \end{tabular}
\end{table}

\subsection{Auxiliary Estimator}
\label{subsec:auxiliary_estimator}
Besides passing the raw sensory information to the policy, we also test passing an auxiliary estimate of the ground truth encoding of the pellet positions $\hat{T}^*$ and of the number of pellets between the fingers $\hat{P}_c$.
A general overview of the estimator is shown in the middle of \cref{fig:architecture}.

The estimator receives a similar input as the policies themselves. 
In particular, we test the estimator for receiving only joint information (policy $\hat{\mathrm{J}}$) and additionally the tactile skin signal (policy $\hat{\mathrm{T}}$).
For the estimator, we use a larger observation stack in contrast to the reinforcement learning policy, hoping to capture more long-term information.
In total, 20 previous observations are stacked, leading to a history of $\SI{1}{\second}$.

First, this stack is passed to a shared feature extractor consisting of two linear layers with 1024 neurons and Swish activation functions.
The features are then passed to a separate position and number head with two hidden layers of 1024 neurons and one layer of 128 neurons, respectively.
As explained in \cref{sec:gt_encoding}, the ground truth encoding $T^*$ can be interpreted as the probability that a contact point falls into a certain area. Therefore, for the position estimate $\hat{T}^*$ we use a binary cross-entropy loss on each individual value.
For the number estimate $\hat{P}_c$, we one-hot encode the ground truth number of pellets between the fingers $P_c$ and use a cross-entropy loss.

The estimator is trained in a supervised fashion alongside the PPO policy.
At every PPO update step, we first collect new data for the estimator training using the latest policy.
The data is added to a buffer that is then used to train the estimator.
This way, the estimator sees a wider variety of inputs avoiding potential overfitting to the latest policy.
For the policy training itself, we then again use the latest version of the estimator.

\subsection{Domain Randomization}
Besides the randomization of the initial grasp, we apply further domain randomization during training to enable a robust sim-to-real transfer.
The most important parameters are explained in the following.
Please visit the accompanying website for an exhaustive list of parameters and their values.
\subsubsection*{Hand Parameters}
For the hand, it is most important to randomize the kinematics as the model is never perfect.
Therefore, we introduce constant measurement offsets in the joint angles. 
Further, to imitate effects like backlash, we add additional Gaussian measurement noise.
Moreover, to make the policy robust against the contact dynamics, we randomize the friction and softness of the fingertips.
\subsubsection*{Skin Parameters}
To allow a successful sim-to-real transfer, the randomization of the tactile skin is also important.
We use a similar randomization as in \cite{kasolowsky2024}.
The most important parameters are the sensor pose, the elasticity and the measurement gains.
The elasticity determines how widely the contact spreads between the taxels.
The measurement gain is a scaling constant that determines the ratio between the force applied within one taxel and the effective measurement.
We further add measurement noise to the signal.

\subsection{Reinforcement Learning Training}
For the implementation, we set up our environment according to the Gymnasium API \cite{brockman2016} and use the PPO implementation of Stable Baselines3 \cite{raffin2021}.
The actor and the critic networks are simple MLPs with two layers, which have 1024 neurons each.
To allow a more stable training of the critic network, we utilize asymmetric observations \cite{akkaya2019,andrychowicz2020}: the critic additionally receives the ground truth encoding of the pellet position and the current number of pellets between the fingers as privileged information.
For data collection, we use 160 Mujoco environments in parallel.
We train each policy for a total of \num{8000000} environment steps.
This takes, without the estimator, approximately \SI{1.5}{\hour} on a machine with 20 CPU cores and an Nvidia T4.
If an estimator is trained in parallel, training time increases to roughly \SI{4}{\hour}.

\section{Results}
In this section, we first analyze the results of learning the given separation task in simulation. 
Moreover, we present the successful transfer from simulation to the real hand in combination with the tactile skin.
Our main metric for evaluation is the success rate $p(P_c=P_d)$ that describes the ratio of how often the desired number of pellets is matched after an episode of \SI{5}{\second}.

\subsection{Analysis in Simulation}
\subsubsection*{Basic Policies without Estimator}
\begin{figure}
    \centering
    \includegraphics[width=0.66\linewidth]{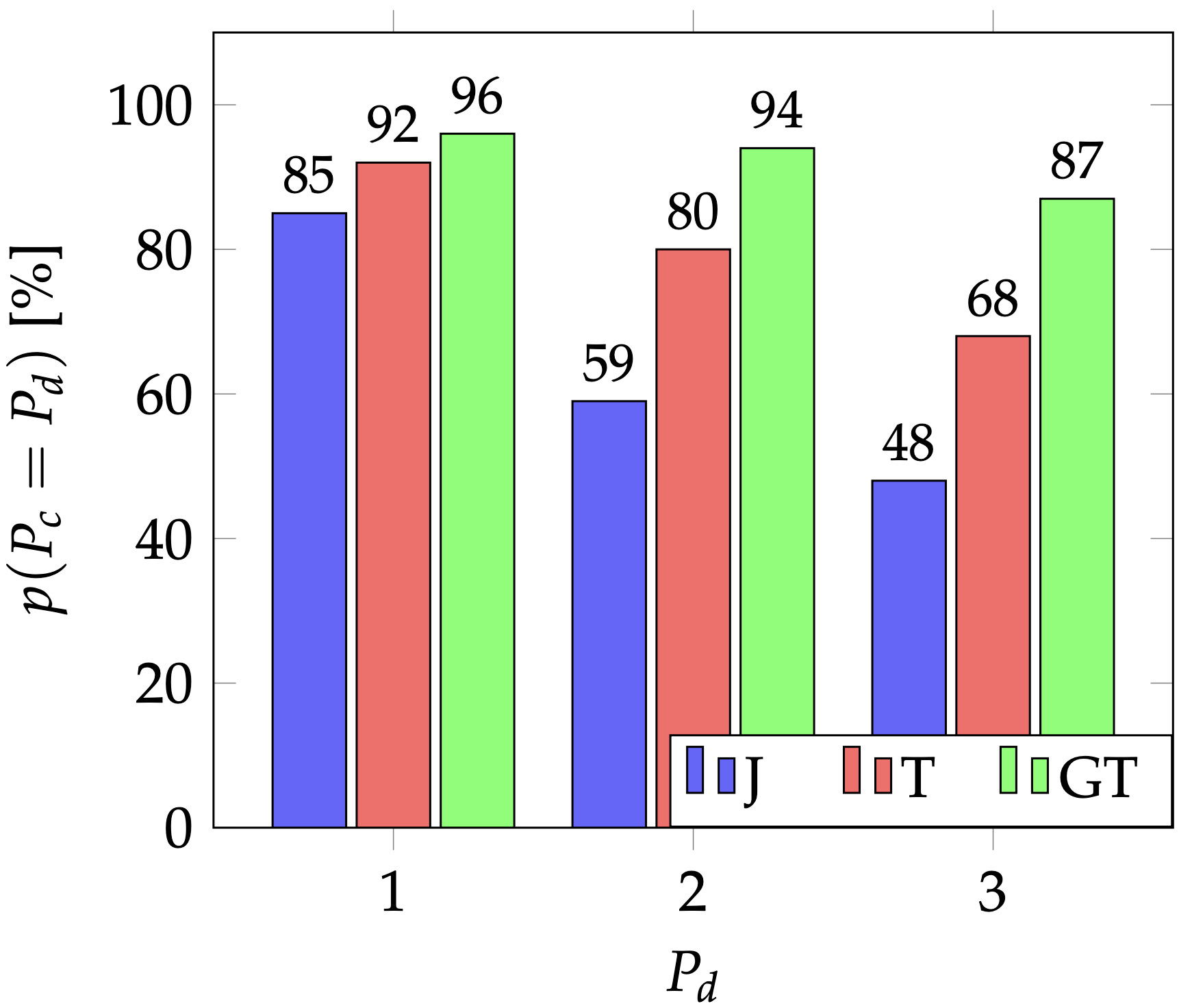}
    \caption{Results of learning in simulation. Success rate $p(P_c=P_d)$ of different observation spaces for different target numbers $P_d$ in simulation.}
    \label{fig:simulation_statistic}
\end{figure}

First, we train the task without the estimator for a desired number of one, two or three pellets: $P_d \in \{1,\, 2,\, 3\}$.
We evaluate the policy for \num{500} runs each.
The resulting success rates are shown in \cref{fig:simulation_statistic}.

As expected, the ground truth policy $\mathrm{T^*}$ performs always the best.
For a desired pellet number of one, it solves the task almost perfectly with a success rate of \SI{96}{\percent}. 
We see that the task gets harder with an increasing desired number of pellets, as the success rate drops to \SI{94}{\percent} for $P_d=2$ and to \SI{87}{\percent} for $P_d=3$.
This indicates that it requires an increasingly sophisticated strategy to decide which pellets should be kept and which should be dropped in what order.

This is supported by results of the policy $\mathrm{J}$ that has no tactile feedback, i.e., it has no access to the pellet positions at all.
For one desired pellet $P_d=1$, it finds a blind strategy achieving a success rate of \SI{85}{\percent}.
If multiple pellets should remain between the fingers, the performance suddenly drops to \SI{59}{\percent} and \SI{48}{\percent} for $P_d=2$ and $P_d=3$, respectively.

Despite having a small resolution of only $4 \times 4$ taxels and a small size that does not cover the entire fingertip, the signal of the tactile skin drastically improves the performance.
The effect is most visible for the harder tasks of $P_d=2$ and $P_d=3$, where the policy $\mathrm{T}$ leads to an increase in the success rate of \SI{21}{\percent} and \SI{20}{\percent}, respectively, compared to policy $\mathrm{J}$.

\subsubsection*{Auxiliary Estimator}
We test the training with an estimator for a desired pellet number of $P_d=2$ first with only the joint angles (policy $\hat{\mathrm{J}}$) and then with feedback of the tactile skin (policy $\hat{\mathrm{T}}$).
The results of both the estimator and the policy are listed in \cref{tab:estimator}.
We see that the estimator improves the final success rate marginally by \SI{4}{\percent}, respectively.
However, it allows for some interesting insights into the effectiveness of the tactile skin.

\begin{figure}
    \centering
    \includegraphics[width=0.9\linewidth]{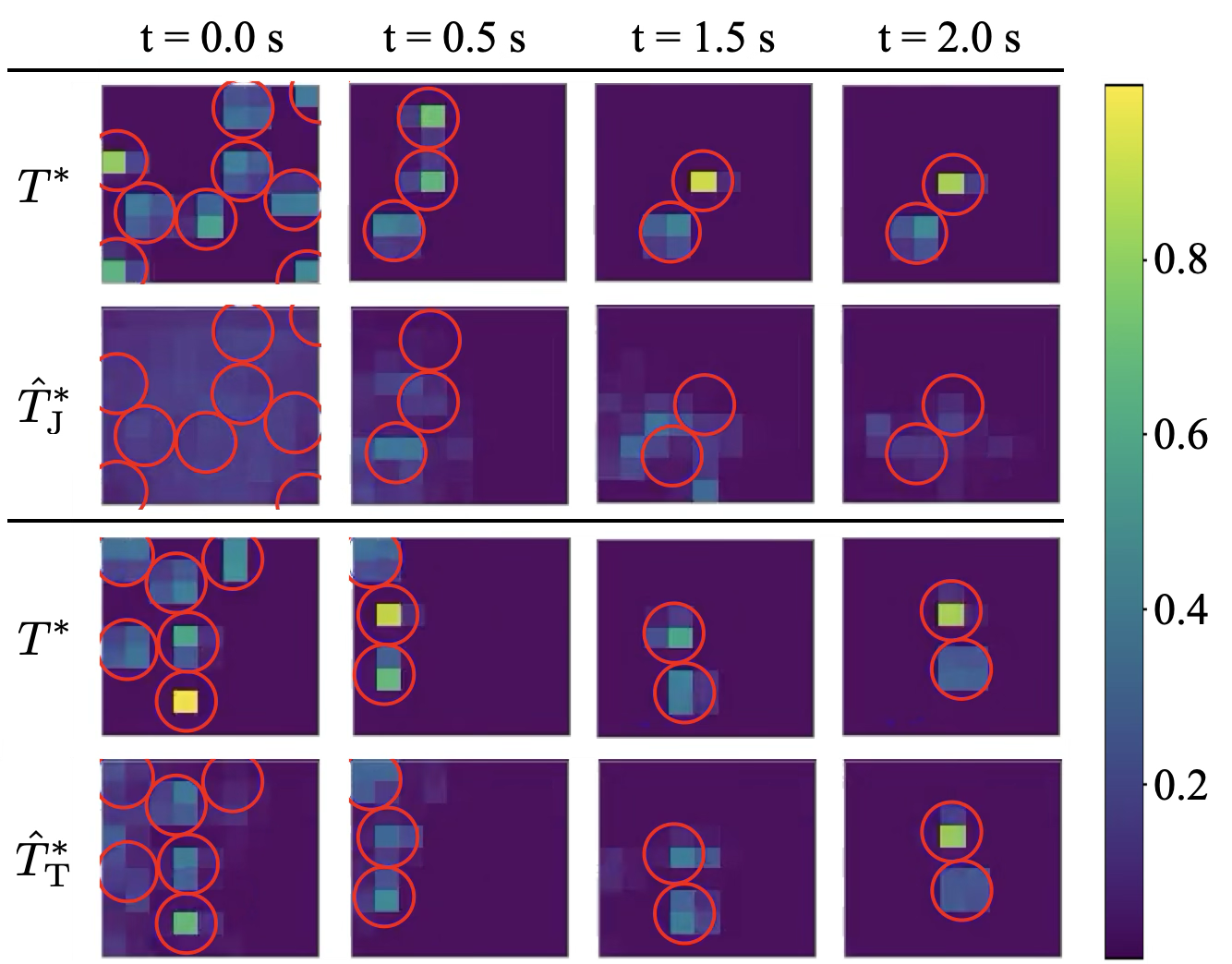}
    \caption{
        Visualization of the position estimate. Example sequences for the estimator without ($\hat{T}^*_\mathrm{J}$) and with tactile feedback ($\hat{T}^*_\mathrm{T}$).
        The red circles indicate the pellets on the fingertip.
        For each run, the top row shows the ground truth encoding and the bottom row shows the reconstruction.
    }
    \label{fig:estimator_position}
\end{figure}

The estimator without tactile feedback achieves a validation accuracy of \SI{60}{\percent} for predicting the correct current number of pellets between the fingers. 
In contrast, with tactile feedback the accuracy is \SI{81}{\percent}.
Looking at the estimate of the pellet positions, the difference becomes more visual.
See \cref{fig:estimator_position} for two example episodes and the corresponding estimates at different time steps.
At the initial time step, for the joint angles the prediction $\hat{T}_{\mathrm{J}}^*$ is just a big cloud on the entire finger.
In contrast, with tactile feedback the estimate $\hat{T}_{\mathrm{T}}^*$ roughly knows where potential pellets could be.
Pellets at the edge of the fingertip are hard to localize even with tactile feedback, as the skin at hand has a smaller coverage.
After the separation is performed, the two remaining pellets can be localized very well with the tactile skin.
Without it, however, the prediction remains blurry.

\begin{table}[!htb]
    \caption{Auxiliary estimator for $P_d=2$.}
    \label{tab:estimator}
    \centering
    \begin{tabular}{lccc}
        \hline
        \multicolumn{2}{c|}{} & $\hat{\mathrm{J}}$ & $\hat{\mathrm{T}}$ \\\hline
        Position Loss & $\mathcal{L}_p$ & $\SI{0.107}{}$ & $\SI{0.074}{}$ \\
        Number Loss & $\mathcal{L}_n$ & $\SI{0.976}{}$ & $\SI{0.543}{}$ \\
        Number Accuracy & $A_n$ & $\SI{60}{\percent}$ & $\SI{81}{\percent}$ \\\hline
        Success rate & $p(P_c=P_d)$ & $\SI{63}{\percent}$ & $\SI{84}{\percent}$
    \end{tabular}
\end{table}

\subsection{Simulation to Reality Transfer}
\begin{figure}
    \centering
    \includegraphics[width=0.66\linewidth]{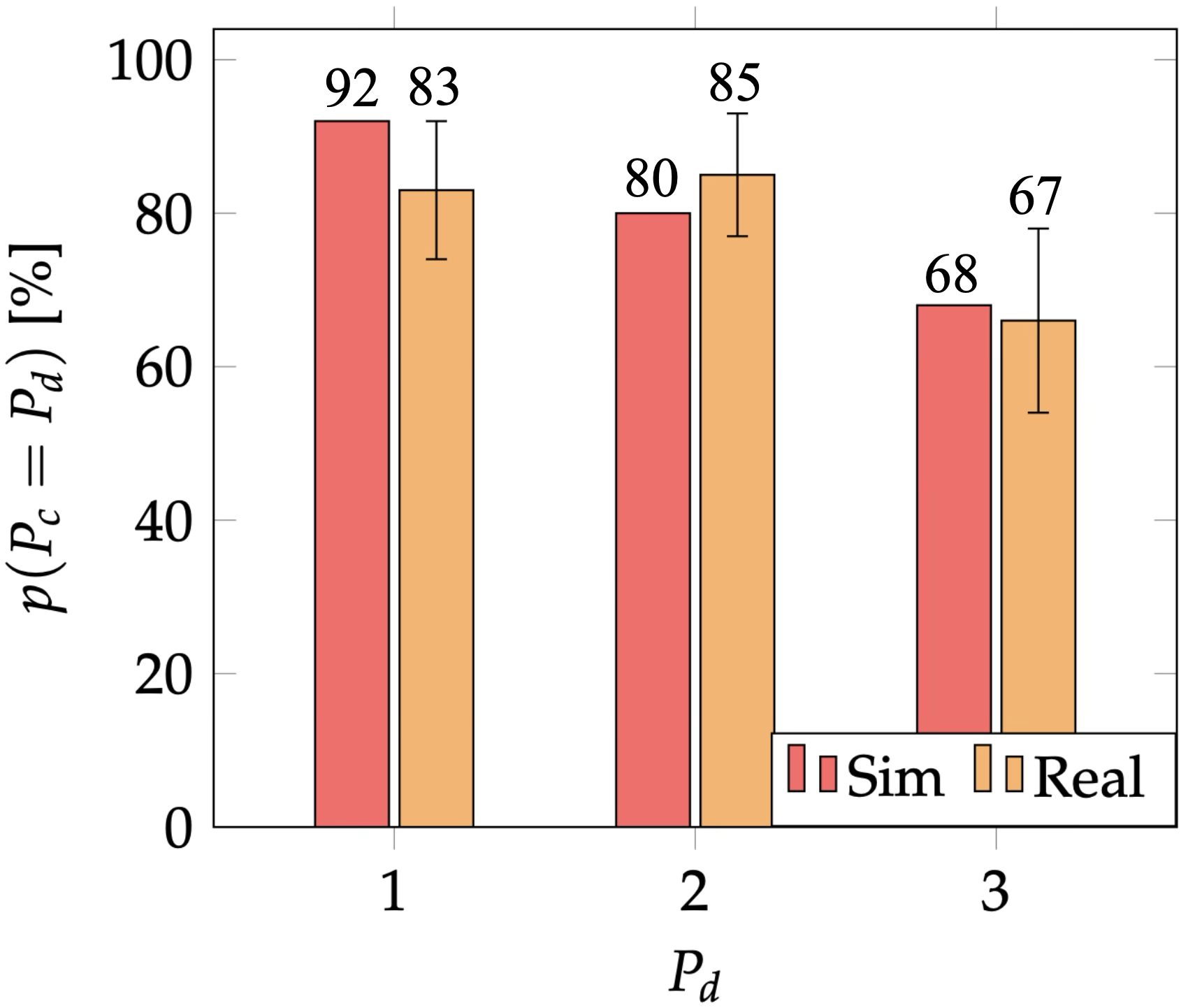}
    \caption{
        Results of the sim-to-real transfer of the tactile policy. 
        Success rate $p(P_c=P_d)$ for different target numbers $P_d$.
        For comparison, we also show the corresponding success rates in simulation.
        For the results in real life, the corresponding $1\sigma$ confidence interval is also shown.
    }
    \label{fig:reality_statistic}
\end{figure}

For the sim-to-real transfer, we test the policy $\mathrm{T}$ with the tactile skin and without an estimator.
For each desired number of pellets, we perform 15 to 20 runs.
The resulting success rates are shown in \cref{fig:reality_statistic} together with the resulting $1\sigma$ confidence interval that is computed as
    $\sigma = \sqrt{\hat{p}*(1-\hat{p})/N}$
where $\hat{p}$ is the success rate and $N$ is the number of trials.
We can observe that the found success rates match well with the ones in simulation, proving a successful sim-to-real transfer.

\begin{figure}
    \centering
    \includegraphics[width=0.66\linewidth]{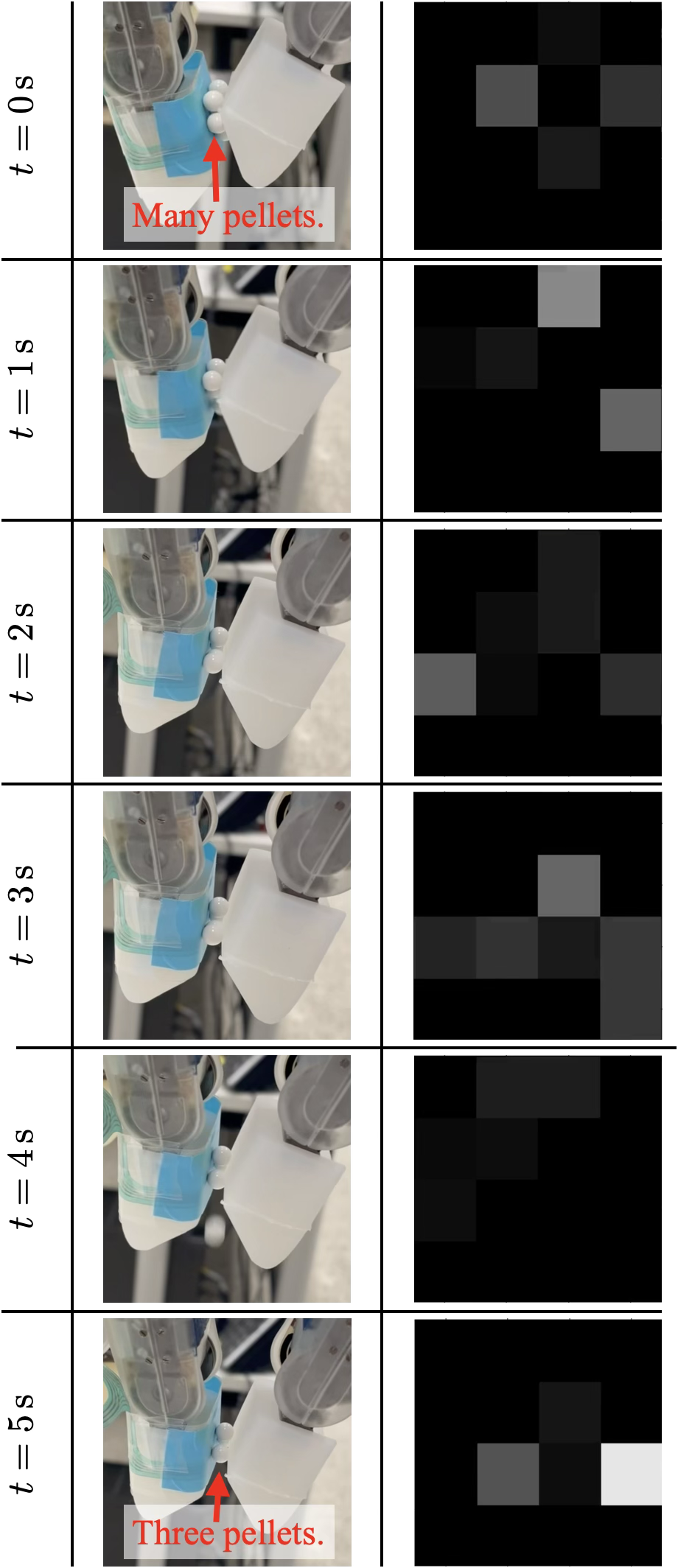}
    \caption{
        Example run on the real system for $P_d=3$ with tactile feedback.
        On the left, a sequence of the manipulation is shown over time. 
        On the right, the corresponding tactile images are visualized (black: no activation, white: high activation).
        Also see the accompanying video to get the best impression.
    }
    \label{fig:sequence_3}
\end{figure}

An example run for $P_d=1$ is illustrated in \cref{fig:title_figure}. 
\Cref{fig:sequence_3} shows a run for $P_d=3$ in more detail, including the observed skin signal.
Furthermore, we advise the reader to watch the complementary video to get the best impression of the sim-to-real transfer.

\section{Conclusion}
We have shown that the task of controlled separation between two fingers of small pellets to a desired number can be learned in simulation and successfully transferred to a real multi-purpose robotic hand.
We performed an exhaustive analysis of the task in simulation.
With a perfect tactile sensor, the task can be solved almost perfectly.
For a desired number of one, two, and three, pellets success rates of \SI{96}{\percent}, \SI{94}{\percent}, and \SI{87}{\percent} were achieved, respectively.

We have further shown that even a tactile sensor with a spatial resolution of only $4\times4$ taxels on one fingertip improves the success rate for the case of multiple desired pellets ($P_d=2$ or $P_d=3$) by as much as \SI{20}{\percent}.
Finally, a successful sim-to-real transfer was performed for all three desired target numbers with real-world success rates close to the ones from the simulation.

To our knowledge, it was the first time that this controlled separation task with a desired number $\ge 1$ was performed with a robotic hand.
Further, it was the first time that such a delicate manipulation task with objects smaller than a centimeter was trained purely in simulation.

For future work, we plan to expand the controlled separation to other objects like screws or nuts, especially posing challenges for the simulation due to more complex contacts and dynamics.

\footnotesize
\bibliographystyle{IEEEtranN-modified}
\bibliography{IEEEabrv, bibliography}

\end{document}